\documentclass[12pt]{article}
\usepackage{graphicx} 
\usepackage{arxiv}
\usepackage{palatino}
\usepackage[super]{natbib}

\graphicspath{ {.} }

\usepackage{enumitem}
\usepackage[most]{tcolorbox}
\usepackage[colorlinks]{hyperref}

\usepackage{gb4e}

\newtcolorbox{abox}[1][]{breakable,pad at break*=1mm,colback=blue!5!white,colframe=blue!75!black, title=#1}

\title{Toward best research practices in AI Psychology}
\author{Anna Ivanova 
\\Georgia Institute of Technology\\
a.ivanova@gatech.edu\\}
\date{July 2024}

\begin{document}

\maketitle


\begin{abstract}
Language models have become an essential part of the burgeoning field of AI Psychology. I discuss 14 methodological considerations that can help design more robust, generalizable studies evaluating the cognitive abilities of language-based AI systems, as well as to accurately interpret the results of these studies.
\end{abstract}


\section*{Cognitive evaluations of AI models}

Ever since the Turing test, the idea of having a dialogue with a machine to probe its cognitive abilities (``thought'') has been inextricably associated with the field of artificial intelligence (AI). In addition to its intuitive simplicity, this idea naturally aligns with everyday practice in psychology: language-based assessments are the bread and butter of many psychologists' toolkits. If researchers want to know what is happening in the mind of a human, the easiest approach is to ask. 

Today, advances in linguistic abilities of large language models (LLMs)---and AI systems that might incorporate LLMs as one of their components---make it possible to seamlessly test these models on language-based assessments originally designed for people. This is an unprecedented advance: to date, the only entities who could flexibly use human language were, well, human. Now, however, we are faced with artificial systems that can process linguistic information, generate novel texts, and respond to questions. How do we assess the cognitive capabilities of these systems?

Easy access to chat-based LLM interfaces (such as ChatGPT) makes it possible for anyone to run a ``cognitive test'' on an AI system. This advance has led to an explosive growth of \emph{AI psychology}, with papers assessing LLMs'  working memory capacity, logical reasoning, planning abilities, social reasoning,  creativity, and even personality traits. Advances in vision-language models now also make it possible to test AI systems on cognitive assessments that incorporate pictures and videos.

Although running such assessments can be fairly straightforward, interpreting the results is not. In fact, AI Psychology today is faced with a plethora of methodological questions. What factors should we consider when assessing model performance? How can we adapt our stimuli to reduce the prevalence of ``hacks'', i.e., heuristics that the model might employ to achieve high task performance without using the cognitive skill being assessed? If a model passes a human test for a cognitive ability $X$, does it mean that it indeed possesses $X$? 

To begin answering these questions, I provide a list of 14 do's and don'ts to consider when designing, conducting, and interpreting the results of an AI Psychology study (see also \citep{frank2023baby} and \citep{mitchell2023we}).
I do not touch on the philosophical question of whether it is at all appropriate to ascribe mental capacities to a machine; my goal here is simply to clarify the methodological criteria that determine the inferences we can(not) make based on a model's responses to a questionnaire or a cognitive test. 

\begin{figure}[htb]
    \centering
    \includegraphics[width=0.9\textwidth]{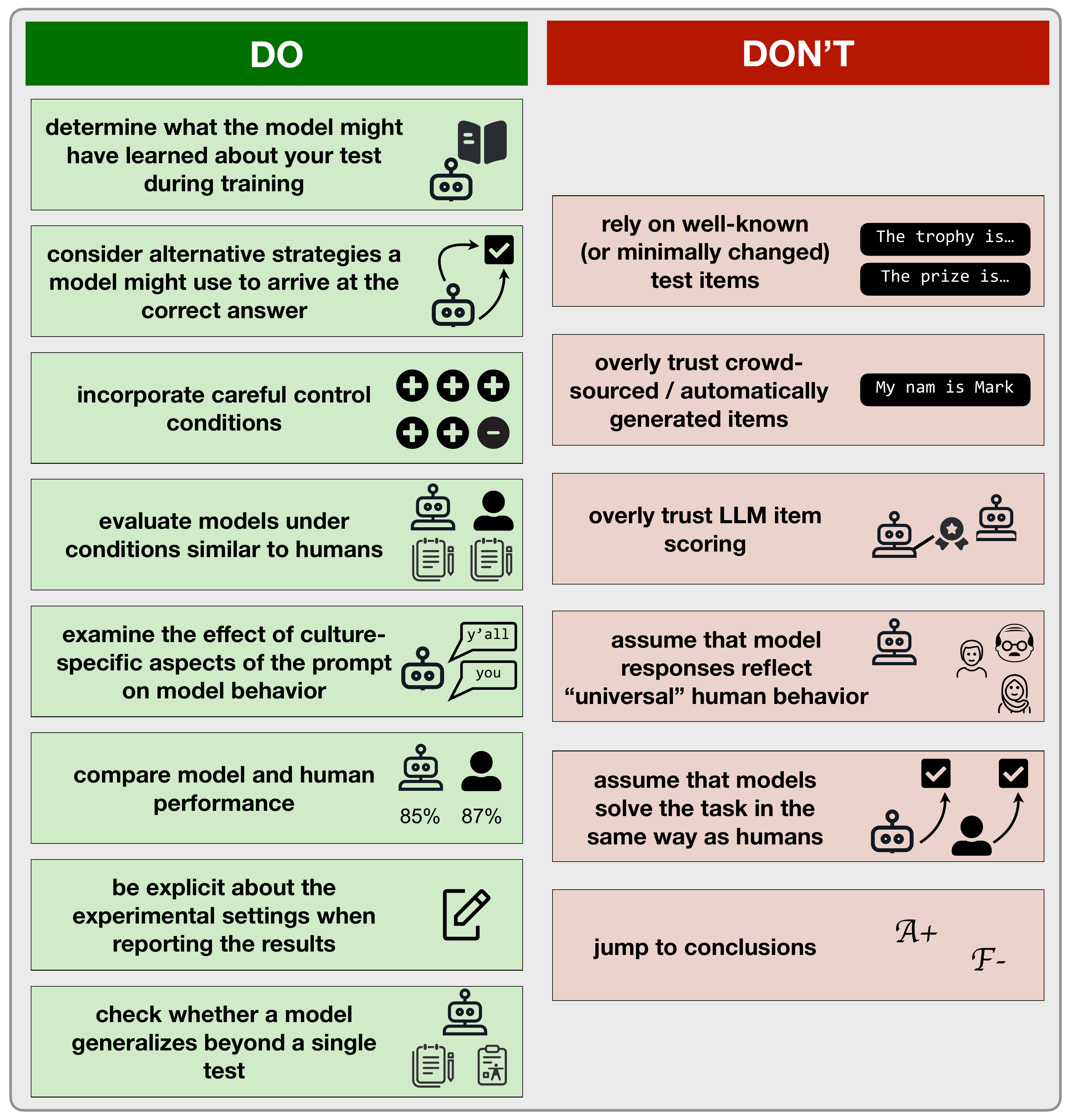}
    \caption{The do's and don'ts of AI model evaluation on cognitive tests.}
    \label{fig:main}
\end{figure}

\section*{Experimental Design} 

When designing or adapting a cognitive test for evaluating an AI model:

\begin{enumerate}

    \item \textbf{DO determine what the model might have learned about your test during training}. The two most important issues to consider are: (a) was the model directly trained on your task? (b) did the model ``see'' examples from your test during its training? In the worst case, \emph{a} and \emph{b} might occur together, i.e., the model was trained on the task of interest using the same items as those in the current study. 

    \item \textbf{DO consider alternative strategies a model might use to arrive at the correct answer}. Even if the model has not directly learned your task during training or finetuning, it might still use a different strategy from the strategy you, the experimenter, have presupposed. The most prominent shortcuts for LLMs include word associations based on their statistical co-occurrence (``racecar''/``fast''), although there might be many others, such as grammatical cues or more abstract structural patterns in text. Humans, too, routinely use task strategies not initially considered by the experimenters, but the kinds of alternative strategies available to humans and to AI systems are not necessarily the same. Considering possible shortcut paths to solutions can help both design shortcut-proof items and identify edge cases where shortcut-based models are likely to fail \citep{mccoy2024embers}. 
    
    \item \textbf{DO incorporate careful control conditions}. 
    Once you identify alternative strategies by which a model might solve the task, the next step is to design control conditions to help distinguish these possible strategies. For instance: did the model choose answer \emph{a} over \emph{b} simply because \emph{a} is more frequent? Will the model show the same preference for answer \emph{a} over \emph{b} even when the question is omitted?
  
    \item \textbf{DO NOT rely on well-known (or minimally changed) test items}. When tested on a famous test item (such as ``The trophy did not fit the suitcase because it was too small''; Box 1), the model has likely seen it multiple times during training and therefore has a high chance of relying on answer memorization; thus, its performance may not generalize to new, non-famous test items. Even if the experimenter makes a minor change to the original item, such as replacing ``trophy'' with ``prize'', a model might still be able to complete the test based on simple association between the old and the new item.
    
    \item \textbf{DO NOT overly trust crowd-sourced / automatically generated items}. The ability of AI models to process large amounts of queries---way more than any human---creates a temptation of evaluating their performance on thousands of test items, obtained from online human workers, created by automatic item generation scripts, or even generated by AI models themselves. However, if the quality of these items is not rigorously evaluated, such tests may be meaningless. AI-generated items present their own set of challenges: a model might (a) generate test items that resemble samples from the training data or, more broadly, (b) generate patterns that have a high likelihood under the model and might therefore facilitate its performance. The systematic assessment of such biases in AI-generated materials remains to be done.
    
    \item \textbf{DO evaluate models under conditions similar to humans when comparing their performance}. When conducting direct model-to-human comparisons, the default should be to recreate the same testing conditions for humans and models \citep{lampinen2022can}. If a human needs instructions to perform well on a test, the model should also receive those instructions. If a human performs well on a task with no in-the-moment training, a model with human-level performance would also need to perform the task with no in-context training. If divergences during evaluation occur (for either theoretical or practical reasons), they need to be justified and highlighted when presenting the results.

    \item \textbf{DO examine the effect of culture-specific aspects of the prompt on model behavior}. The language in which the test is presented, the dialect, the vocabulary used, and even the use of capitalization and punctuation might all affect AI model performance as it leverages these cues to mimic specific users online. 

\end{enumerate}

\section*{Interpreting the results} 

When interpreting the results of a cognitive assessment:

\begin{enumerate}[resume]

    \item \textbf{DO compare model and human performance}. We often assume humans will perform well on a specific test even when we don't have direct evidence for it. If these exact test items and question/prompt wordings have not been tested on humans before, they should be. 

    \item \textbf{DO be explicit about the experimental settings when reporting the results}. Computational work should, ideally, have perfect replicability. For that to be achievable, all the materials and code needed to run the experiment should be made available online. For models offered by commercial third parties, there is no guarantee that they will continue to offer access to the model, so some backup arrangements may need to be made. Finally, a third party may roll out a model update that would change experimental results; in such cases, the date of the experiment needs to be reported as well.

    \item \textbf{DO NOT overly trust LLM item scoring}. The flip side of testing a model on thousands of items is the need to evaluate responses to all these items. In some cases, determining correct answers is straightforward (e.g., a-d for a multiple-choice question); however, scoring free response items is hard work. Many studies are opting for LLM-based item evaluation; however, such approaches might be circular because the items that are hard for the model being evaluated might also be hard for the model doing the evaluation. At a minimum, the model being tested and the model doing the evaluation should belong to different classes; and even then, evaluator model performance needs to be rigorously checked by comparing it with human evaluations across a range of item difficulties.

    \item \textbf{DO NOT assume that model responses reflect ``universal'' human behavior}. Even when directly comparing AI models and humans, it is important to be specific about which human population serves as a reference. Human psychology suffers from over-reliance on WEIRD (white, educated, industrialized, rich, democratic) participant pools; AI Psychology suffers from similar issues \citep{atari2023humans} because WEIRD individuals disproportionately contribute to the training data. Thus, calls to replace human data in psychology studies with AI-generated responses need to account for the strong demographic and cultural biases that these models bring. Similarly, model performance in English (the most represented language on the web) might not be reflective of model performance in other languages.

    \item \textbf{DO NOT assume that models solve the task in the same way as humans}. Even if both humans and models do well on a particular test, it does not mean that they solve it in the same way. Some approaches to evaluating similarities and differences in the mechanisms underlying human and model task performance include comparing their error patterns,  generalization to new tasks, and the internal (neural) representations required to accomplish the task.

    \item \textbf{DO check whether a model generalizes beyond a single test}. Even when we control for possible shortcuts, a model might still find a loophole that allows it to perform well on a particular test. However, the more diverse tests we include (and the more we control for the shortcuts in each), the harder it will be to attribute model performance to serendipitous factors. If a model does well on 20 logical reasoning tasks that vary in their content and format, it is more plausible that it possesses logical reasoning than if it performs well on one such task.

    \item \textbf{DO NOT jump to conclusions}. The discourse around AI models has become very polarized, with both extreme enthusiasm based on a few isolated examples and extreme skepticism based on isolated failures. What the field needs is careful evaluation of specific model capabilities with a clear acknowledgement of advances and limitations. Moreover, results reported for one model may not generalize to other models, so a single study's conclusions should be qualified accordingly.

\end{enumerate}




\begin{abox}[Case study: Winograd Schema Challenge] \label{box-winograd}

In 2012, Levesque and colleagues proposed a test for diverse kinds of basic world knowledge \citep{levesque2012winograd} (named after an initial example by Terry Winograd). The test includes minimally different item pairs with pronouns whose referent can be inferred based on general world knowledge:

\begin{exe}
	\ex  Q: The trophy doesn't fit into the brown suitcase because it's too small. What is too small? \\ A: The suitcase
    \ex  Q: The trophy doesn't fit into the brown suitcase because it's too large. What is too large? \\ A: The trophy
\end{exe}


A set of such minimally differing sentence pairs, tackling diverse world knowledge phenomena --- physical properties, biology, social situations, etc., --- was then compiled into the Winograd Schema Challenge (WSC). Although initially challenging, the WSC was largely solved by 2020, with LLMs achieving over 90\% accuracy \citep{sakaguchi2021winogrande}. However, it turned out that this success did not reflect models' deep knowledge of the world but rather the flaws in the test itself. 

\vspace{6pt}

In the original paper introducing the challenge \citep{levesque2012winograd}, the authors cautioned against examples that could be solved via simple heuristics. They cite two such heuristics: selectional restrictions and frequency effects. The selectional restrictions heuristics can be illustrated with the example: ``The women stopped taking the pills because they were pregnant/carcinogenic''. Here, only an animate entity can be pregnant and only an inanimate entity can be carcinogenic, leading to an unambiguous association between the adjectives and the corresponding nouns without the need to tap into deeper world knowledge. The frequency heuristics applies to cases like: ``The racecar zoomed by the school bus because it was going so fast/slow'': a simple association between ``racecar'' and ``fast'' will suffice to determine the referent. Finally, the authors note that the examples need to be, in their words, \emph{Google-proof}, i.e., not present in online text corpora used to train the models.

\vspace{6pt}

Despite the field's awareness of these heuristics, constructing a heuristics-free dataset was hard. In the original WSC dataset, over 10\% of the items turned out to have simple association-based solutions \citep{trichelair-etal-2019-reasonable}. To address this issue, Sakaguchi et al \citep{sakaguchi2021winogrande} constructed a large dataset called WinoGrande, where the examples were first crowd-sourced online and then automatically filtered to reduce co-occurrence-based associations as estimated through language model embeddings (rather than human intuition). This filtering substantially reduced model performance, suggesting that the association heuristic indeed inflates model accuracy. 

\vspace{6pt}

Soon after, Elazar et al \citep{elazar-etal-2021-back} showed that adding even more stringent quality controls leads to significant decreases in model performance on the WSC. 
The authors introduced two baseline cases to account for raw frequency of possible continuations: sentences with candidate referents excluded (``doesn't fit into because it's too large/small'') and sentences with the first half excluded (``because it's too large/small''). The assumption is that for these reduced sentences, there should be no consistent preference for ``suitcase'' vs. ``trophy'', and if there is, it reflects an association bias. It turned out that LLMs performed above chance on the WSC even in those baseline conditions, indicating previously undetected association biases. 

\vspace{6pt}

The final issue raised by the WSC story is the quality vs.~quantity tradeoff in test design. The initial WSC set includes less than 300 examples, all hand-crafted by scholars. Now that these examples are freely available online, performance on this set is no longer reflective of the models' general capabilities. The authors of WinoGrande used a different approach: to obtain thousands of novel items, they asked human workers to come up with many different examples and then used automatic filtering techniques. This approach is more scalable but suffers from numerous quality issues, e.g. typos (``wit'' instead of ``with''), grammatical errors (``more brighter color''), and unwarranted assumptions (``good at math'' means ``likely to be a professor''). Overall, the tradeoff between result generalizability (which is harder for small datasets) and quality control (which is harder for large datasets) remains an important issue to consider in test design.

\vspace{6pt}

Kocijan et al \citep{kocijan2023defeat} formulate several lessons from the WSC saga, the most important of which is perhaps: ``We need to be careful not to rely on a perceived connection between tasks and methods''. Just because we think that a certain task requires a cognitive ability X, doesn't mean that it actually does.

\end{abox}


\section*{Testing open vs. closed models}

Some of the \textsc{do's} discussed above --- checking the training data for contamination with test items, verifying that the model has not been fine-tuned on the exact task being tested, creating conditions that will allow the study to be replicated in the future on the exact same model --- are essentially impossible in the case of closed models. 
Thus, there is an argument to stop running AI Psychology assessments on closed models altogether given that, from a scientific perspective, the results are potentially non-reproducible and difficult to interpret. 

Will researchers indeed stop running cognitive tests on closed AI models? Probably not. Although scientifically questionable, evaluating the behavior of closed, industry-standard models has important practical implications, including understanding which real-life tasks they can (and cannot) be reliably used for, what AI models are in principle capable of achieving (given that closed models often exhibit state-of-the-art performance), and what safety risks they may present. Thus, going forward, cognitive evaluations of AI systems might be used in two separate settings: (a) basic scientific inquiry of cognitive capacities of AI systems --- which should prioritize open models, and (b) applied studies tackling questions related to model performance, safety, and user impact --- which can be applied both to open and to closed models with the caveat that closed model results might not replicate or generalize.



As with any rapidly growing research area, the scientific practices and norms in AI Psychology get defined on the fly, often through trial and error. I hope that this discussion of the \textsc{do's} and \textsc{don'ts} of AI Psychology will help distill some of the lessons learned and improve the robustness and validity of future work aiming to probe the cognitive capacities of AI systems.


\section*{Acknowledgements}

I gratefully acknowledge the funding support from the University System of Georgia. Many thanks to Aalok Sathe, Ben Lipkin, Carina Kauf, Greta Tuckute,  Kyle Mahowald, and the reviewers for their constructive comments.

\bibliographystyle{naturemag}  
\bibliography{references,anthology} 

\begin{thebibliography}{10}
\expandafter\ifx\csname url\endcsname\relax
  \def\url#1{\texttt{#1}}\fi
\expandafter\ifx\csname urlprefix\endcsname\relax\def\urlprefix{URL }\fi
\providecommand{\bibinfo}[2]{#2}
\providecommand{\eprint}[2][]{\url{#2}}

\bibitem{frank2023baby}
\bibinfo{author}{Frank, M.~C.}
\newblock \bibinfo{title}{Baby steps in evaluating the capacities of large language models}.
\newblock \emph{\bibinfo{journal}{Nature Reviews Psychology}} \textbf{\bibinfo{volume}{2}}, \bibinfo{pages}{451--452} (\bibinfo{year}{2023}).

\bibitem{mitchell2023we}
\bibinfo{author}{Mitchell, M.}
\newblock \bibinfo{title}{How do we know how smart {AI} systems are?} (\bibinfo{year}{2023}).

\bibitem{mccoy2024embers}
\bibinfo{author}{McCoy, R.~T.}, \bibinfo{author}{Yao, S.}, \bibinfo{author}{Friedman, D.}, \bibinfo{author}{Hardy, M.~D.} \& \bibinfo{author}{Griffiths, T.~L.}
\newblock \bibinfo{title}{Embers of autoregression show how large language models are shaped by the problem they are trained to solve}.
\newblock \emph{\bibinfo{journal}{Proceedings of the National Academy of Sciences}} \textbf{\bibinfo{volume}{121}}, \bibinfo{pages}{e2322420121} (\bibinfo{year}{2024}).

\bibitem{lampinen2022can}
\bibinfo{author}{Lampinen, A.~K.}
\newblock \bibinfo{title}{Can language models handle recursively nested grammatical structures? a case study on comparing models and humans}.
\newblock \emph{\bibinfo{journal}{arXiv preprint arXiv:2210.15303}}  (\bibinfo{year}{2022}).

\bibitem{atari2023humans}
\bibinfo{author}{Atari, M.}, \bibinfo{author}{Xue, M.~J.}, \bibinfo{author}{Park, P.~S.}, \bibinfo{author}{Blasi, D.} \& \bibinfo{author}{Henrich, J.}
\newblock \bibinfo{title}{Which humans?}
\newblock \emph{\bibinfo{journal}{PsyArXiv}}  (\bibinfo{year}{2023}).

\bibitem{levesque2012winograd}
\bibinfo{author}{Levesque, H.}, \bibinfo{author}{Davis, E.} \& \bibinfo{author}{Morgenstern, L.}
\newblock \bibinfo{title}{The {Winograd Schema Challenge}}.
\newblock In \emph{\bibinfo{booktitle}{Thirteenth international conference on the principles of knowledge representation and reasoning}} (\bibinfo{year}{2012}).

\bibitem{sakaguchi2021winogrande}
\bibinfo{author}{Sakaguchi, K.}, \bibinfo{author}{Bras, R.~L.}, \bibinfo{author}{Bhagavatula, C.} \& \bibinfo{author}{Choi, Y.}
\newblock \bibinfo{title}{Winogrande: An adversarial winograd schema challenge at scale}.
\newblock \emph{\bibinfo{journal}{Communications of the ACM}} \textbf{\bibinfo{volume}{64}}, \bibinfo{pages}{99--106} (\bibinfo{year}{2021}).

\bibitem{trichelair-etal-2019-reasonable}
\bibinfo{author}{Trichelair, P.}, \bibinfo{author}{Emami, A.}, \bibinfo{author}{Trischler, A.}, \bibinfo{author}{Suleman, K.} \& \bibinfo{author}{Cheung, J. C.~K.}
\newblock \bibinfo{title}{How reasonable are common-sense reasoning tasks: A case-study on the {W}inograd schema challenge and {SWAG}}.
\newblock In \emph{\bibinfo{booktitle}{Proceedings of the 2019 Conference on Empirical Methods in Natural Language Processing and the 9th International Joint Conference on Natural Language Processing (EMNLP-IJCNLP)}}, \bibinfo{pages}{3382--3387} (\bibinfo{publisher}{Association for Computational Linguistics}, \bibinfo{address}{Hong Kong, China}, \bibinfo{year}{2019}).
\newblock \urlprefix\url{https://aclanthology.org/D19-1335}.

\bibitem{elazar-etal-2021-back}
\bibinfo{author}{Elazar, Y.}, \bibinfo{author}{Zhang, H.}, \bibinfo{author}{Goldberg, Y.} \& \bibinfo{author}{Roth, D.}
\newblock \bibinfo{title}{Back to square one: Artifact detection, training and commonsense disentanglement in the {W}inograd schema}.
\newblock In \emph{\bibinfo{booktitle}{Proceedings of the 2021 Conference on Empirical Methods in Natural Language Processing}}, \bibinfo{pages}{10486--10500} (\bibinfo{publisher}{Association for Computational Linguistics}, \bibinfo{address}{Online and Punta Cana, Dominican Republic}, \bibinfo{year}{2021}).
\newblock \urlprefix\url{https://aclanthology.org/2021.emnlp-main.819}.

\bibitem{kocijan2023defeat}
\bibinfo{author}{Kocijan, V.}, \bibinfo{author}{Davis, E.}, \bibinfo{author}{Lukasiewicz, T.}, \bibinfo{author}{Marcus, G.} \& \bibinfo{author}{Morgenstern, L.}
\newblock \bibinfo{title}{The defeat of the winograd schema challenge}.
\newblock \emph{\bibinfo{journal}{Artificial Intelligence}} \bibinfo{pages}{103971} (\bibinfo{year}{2023}).

\end{thebibliography}


\section*{Competing interests}
The author declares no competing interests.

\end{document}